\documentstyle[times,psfig]{oldllncs}

\newfont{\sfi}{cmssi8}
\newfont{\sff}{cmss8}

\begin{document}

\def \email {\normalsize}
\def \www {\normalsize}
\def \prog {\small \sf}

\def \gp {\sff}
\def \gpitalic {\sfi}

\title{Character Design for Soccer Commentary}
\author{
  \mbox{
  \parbox[t]{1.3in}{\begin{center}Kim Binsted\\{\small kimb@csl.sony.co.jp}\end{center}}
  }
}

\institute{
  Sony Computer Science Lab\\
  3-14-13 Higashigotanda\\
  Shinagawa-ku, Tokyo 141
  }
\maketitle

\begin{abstract}

In this paper we present early work on an animated talking head
commentary system called {\bf Byrne}\footnote{David Byrne is the lead
singer of the Talking Heads.}. The goal of this project is to develop
a system which can take the output from the RoboCup soccer simulator,
and generate appropriate affective speech and facial expressions,
based on the character's personality, emotional state, and the state
of play. Here we describe a system which takes pre-analysed simulator
output as input, and which generates text marked-up for use by a
speech generator and a face animation system. We make heavy use of
inter-system standards, so that future versions of Byrne will be able
to take advantage of advances in the technologies that it
incorporates.

\end{abstract}

\section{Introduction}

Many natural systems have behaviour complex enough that people will
tend to ascribe personalities to them, and use those personalities as
flawed but powerfully predictive tools. For example, we might summarize a
dog's behavioral tendencies as ``eager to please'' or ``yappy and
spoiled'', and use this assigned personality to predict its future
behaviour. 

Designed characters --- such as characters in films or novels ---
exploit this tendency, expressing their personality so as to
manipulate the observer's expectations to the designer's ends. For
example, the villain in a novel might sneer and speak in a menacing
voice, cueing us to expect villainous behaviour. This expectation
might be reinforced by more static characteristics, such as a strong
Transylvanian accent or a cheekbone scar. Consistency between
expression and action, and also between modalities of expression,
contributes to a character's believability. Believability, in turn,
contributes to the expected predictive value of the character's
perceived personality.

We are interested in the relationship between consistency and
believability, and between expression and perceived personality.  To
explore these issues, we are developing a talking head system which
will generate entertaining, believable commentary on RoboCup
simulator league games \cite{RoboCup}, complete with facial expressions
and affective speech, in (close to) real time.

A goal of this research is to develop an architecture which allows
artists (whose technological skills may vary) to design expressive,
believable characters (in the first instance, talking heads). The
emphasis is on the expression, not the content --- we assume the
pre-linguistic raw content is generated by another system, specialized
to the application. In the context of a particular character, our
system will:

\begin{itemize} 

\item generate appropriate surface natural language text to express that content,

\item transform that text into natural affective speech, and

\item control a face animation, so that appropriate lip movements and facial expressions are generated.
\end{itemize}

In this early stage of the project, we are mostly interested in the
speech and facial animation components of the system.  

The key issue here is {\em appropriateness}. How do we ensure that the
behaviour of the system is appropriate for the designed character in
the given situation? For example, when reporting a scored goal, the
language and facial expressions used might depend strongly on which
team the character supports. Moreover, how do we ensure that the
designed character is appropriate for the application?  A character
which might be perfect for soccer commentary might not be right for,
say, a medical advisory system. 

This begs the question: is a talking head appropriate for soccer
commentary at all? After all, the heads of human sports commentators
are rarely seen on screen during the main action of the game. In fact,
our attraction to this domain is due more to its usefulness for our
research (please see Section~\ref{domain}) than because soccer `needs'
talking heads in any way. Nonetheless, we do believe that an
expressive entertaining talking head commentary would add to the fun
of watching (or in the case of of a video game, playing) soccer. This
has yet to be seen, of course.

\section{Related work}

\subsection{Believable characters}

Recently there has been a great deal of interest in the design and
implementation of characters with personality. For example, the
Virtual Theatre Project at Stanford \cite{HayesRoth2} is working on a
number of animated virtual actors for directed improvisation, basing
their efforts on Keith Johnstone's theories of improvisational theatre
\cite{Johnstone}. They make use of character animations developed as
part of the {IMPROV} project \cite{Improv}, which can take fairly
high-level movement directions and carry them out in a natural,
expressive manner. Related work on agent action selection and
animation has been done as part of the ALIVE \cite{Blumberg} and OZ
projects \cite{Loyall} \cite{Reilly}.

Although these projects have similar goals and assumptions to ours,
our approach differs from theirs on several points. First, our focus
on talking heads (rather than fully embodied agents in virtual
environments) leads to a stronger emphasis on language-related
behaviours. Also, we do not attempt to have the personality of the
character control content selection, or any other action of the agent,
for that matter. Although this sharp distinction between content and
expression might negatively affect the consistency of the character, a
clear separation between content and expression allows the character
to be portable across content generators. For example, you could have
essentially the same character (an aggressive older Scottish man, for
example) commentate your soccer games and read your maps. In the first
case, the content generating application is the RoboCup soccer
simulator, and in the second case it is an in-car navigation system
--- but the character remains the same.

We also do not make a great effort to make the emotion component of
our system cognitively plausible. Designed characters, such as
characters in novels or films, are generally both simpler and more
exaggerated than natural personalities, such as those we perceive in
each other. The goal is not to develop a psychologically realistic
personality, but to generate a consistent and entertaining character
for some application. For this reason, what psychology there is in
Byrne is more folk psychology than modern cognitive science.

\subsection{Game analysis and commentary}
\label{related}

There are at least two existing systems which generate analysis and
commentary for the RoboCup simulation league: MIKE~\cite{MIKE} and
Rocco~\cite{Rocco}.

Rocco is a system for analysing simulation league games and generating
multimedia presentations of games. Its output is a combination of
spoken natural language utterances and a 3-D visualization of the game
itself. The generated language has appropriate verbosity, floridity,
specificity, formality and bias for the game context. It uses a
text-to-speech system to synthesize the spoken utterances.

MIKE is a system developed at ETL which, given raw simulation data as
input, analyses the state and events of the game, chooses relevant
comments to make, and generates natural language commentary in real
time and in a range of languages. It also uses a text-to-speech
synthesizer to generate spoken output.

Our approach differs from the above in that we do not, at present, do
any game analysis at all; instead, we assume pre-analysed game
information as input. This is because our emphasis is on expression,
rather than utterance content. To our knowledge, neither MIKE nor
Rocco attempt to generate affective speech, and neither use face animation.

\section{Enabling technologies}

In this section we outline some of the existing techologies that Byrne
uses, and discuss some issues related to inter-system
standards.

\subsection{The RoboCup simulation league}
\label{domain}

The simulation league of RoboCup~\cite{RoboCup} \cite{SoccerServer}
features teams of autonomous players in a simulated environment. This
is an interesting domain for several reasons. There is no direct
interaction with a user, which simplifies things a great deal --- no
natural language understanding is necessary, for example. More
importantly, having direct access to the state of the simulation
simplifies perception. If our system had to commentate a real
(non-simulated) game, vision and event-recognition issues arise, which
we'd rather avoid.

However, there are some problems with using the output of the RoboCup
simulator directly:

\begin{itemize}

\item The output of the simulator is at a much lower level of
description than that typically used by a football commentator.

\item The simulator has no sense of which actions and states of play
are relevant or interesting enough to be worthy of comment.

\end{itemize}

For this reason, some kind of game-analysis system is a necessary
intermediary between the soccer simulator and Byrne. Either of the
systems described in Section~\ref{related} would be probably be
adequate for this purpose. However, for the present, we rely on
pre-analysed game transcripts.

\subsection{Mark-up languages}

SGML-based~\cite{SGML} mark-up languages are a useful tool for for
controlling the presentation of information at a system-independent
level. Here we make use of three different mark-up languages: one to
indicate the linguistic structure of the text, one to determine how
the text is to be spoken, and one to control the facial animation.

\subsubsection{Global Document Annotation}

Global Document Annotation (GDA) is an SGML-based
standard for indicating part of speech (POS), syntactic, semantic and
pragmatic structure in text \cite{Nagao}. Although our main interest
in this work is emotional expression, this must be underlaid by
linguistically appropriate intonation and facial gestures. Moreover,
the quality of speech synthesis is generally improved by the inclusion
of simple phrase structure and part of speech infomormation, which can
be encoded in GDA.

\subsubsection{Sable}

SABLE~\cite{Sable} is a SGML-based text-to-speech mark-up system
developed by an international consortium of speech researchers.  It is
based on several earlier attempts to develop a standard, namely
SSML~\cite{SSML}, STML~\cite{STML} and JSML~\cite{JSML}. The goal is
to have a system- and theory-independent standard for text mark-up, so
that non-experts in speech synthesis can mark up text in an intuitive
manner which produces reasonable output from a variety of systems.

Although SABLE is in its early stages and does not yet have the
richness required for the generation of affective speech (such as that
described in \cite{Cahn}), it is a useful standard. Here we have
supplemented Sable with some system-specific tags useful for our
purposes. We hope that similar tags will be adopted into the SABLE set
in the future.

\subsubsection{FACS}

FACS \cite{Ekman} stands for the Facial Action Coding System, and is a
set of all the Action Units (AUs) which can be performed by the human
face. It is often used as a way of coding the articulation of a
facial animation ~\cite{Parke}. For example, AU9 is known as the ``nose
wrinkler'', and is based on the facial muscles Levator Labii
Superioris and Alaeque Nasi. There are 46 AUs. 

Although FACS is not an SGML-based mark-up language, it can be
trivially converted into one by treating each AU as empty SGML
element. Here we refer to such a mark-up as FACSML.

\section{Character architecture}

\begin{figure}[t]
\begin{center}
\mbox{\psfig{figure=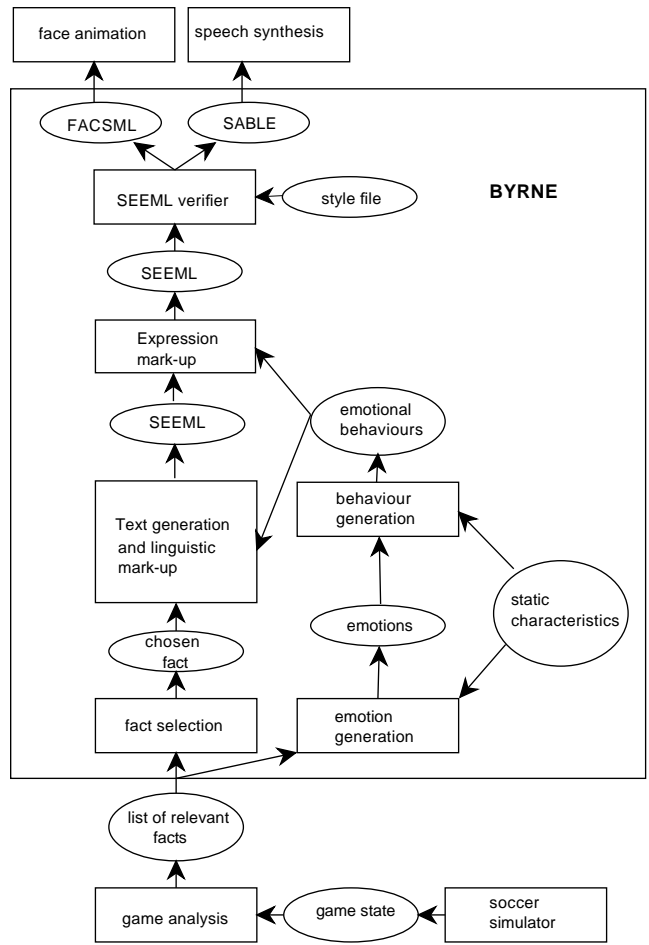,height=5in}}
\end{center}
\caption{The Byrne system architecture}
\label{arch}
\end{figure}

Byrne is a system for expressive, entertaining commentary on a RoboCup
Simulation League soccer game. Here we describe the Byrne character
architecture (see Figure~\ref{arch}).

\subsection{Input}

Byrne's input is a prioritized list of relevant facts about the
current state of and recent actions in the simulated soccer game. We
assume that each fact has been given a relevance score from 1 to
10. For example, when Team A is in possession of the ball and near
Team B's goal, two minutes into the game, player A1 has just passed
the ball to player A2, and player B1 (Team B's `goalie', perhaps) has
just moved towards the ball, the list might read:

\begin{itemize}

\item (pass from: a1 to: a2 fromloc: (30 10) toloc: (20 10) begintime: 120 endtime: 125 relevance: 10)
\item (has-ball player: a2 location: (20 10) relevance: 5)
\item (move player: b1 fromloc: (5 10) toloc: (10 10) begintime: 115 endtime: 120 relevance: 3)
\item \ldots
\end{itemize}

This list is updated every clock tick, with new events added,
relevance scores updated, and events with relevance less than one
removed from the list.

Note that we assume that some analysis has already taken place, and
that significant events and their relative relevance have already been
identified. We also assume that relevance decays with time, although
the assessment of relevance happens in the game analysis system, not
within Byrne.

\label{choice}
In the fact selection module, Byrne chooses the most relevant fact to
be reported. In future, we hope to have the character's biases
(e.g. what team she supports, her favorite player, etc.), but for now
Byrne simply selects the fact with the highest relevance score.

If a fact with higher relevance appears on the list while a fact with
lower relevance is being reported, then the lower relevance utterance
is interrupted at the next interrupt marker (see
section~\ref{markup}), the utterance generated to report the higher
relevance fact is started.

The assumption that the game analysis and relevance calculation has
already taken place is a big one, and puts a large load on the other
system which would do this work. For example, Byrne's interruption
mechanism depends completely on appropriate relevance scores being
provided and maintained. For Byrne's first prototype, we will avoid
these issues by using pre-prepared input, based on Simulation League
logs.

\subsection{Emotion generation}

The emotion generation module contains rules which generate simple
{\em emotional structures}. These structures consist of:

\begin{itemize}

\item {\bf a type}, e.g. {\em happiness, sadness}, etc. At present we
are using Ekman's six basic emotions ({\em fear, anger, sadness,
happiness, disgust} and {\em surprise}) \cite{Ekman}, as research has
shown that these are clearly and unambiguously expressible. We also
include {\em interest}~\footnote{For our purposes, ``interest'' is
that emotion which at high intensity is called ``excitement'' and low
intensity is called ``boredom''.}, as it is important in our domain of
sports reporting, and is relatively easy to express in speech. In the
future, however, we plan to have a hierarchy of emotions available, to
allow for a richer range of resultant behaviours.

\item {\bf an intensity}, scored from 1 to 10. An emotion with
intensity less than one is considered inactive, and is deleted
from the emotion pool. Note that apparently opposite emotions (such as
{\em happiness} and {\em sadness}) are not considered to be true
opposites in Byrne. That is, a {\em sadness} structure is not merely a
{\em happiness} structure with an intensity of -10. This is because {\em
happiness} and {\em sadness} are not opposites in the sense that they
cannot coexist, but only in that the behaviours they tend to inspire
often conflict --- that is, it is hard to express joy and sadness at
the same time. Thus, this apparent conflict is resolved in the
emotional behaviour rules, rather than in the emotional structures
themselves. Moreover,emotion structures of the same type, but with
different causes and/or targets, can coexist in the emotion pool.

\item {\bf a target} [optional]. Some emotions, such as {\em anger}
and {\em interest}, are usually directed at some person or
object. This is the {\bf target} of that emotion.

\item {\bf a cause}. This is the fact about the world (in this case,
the soccer game) which caused the emotion to come into being. 

\item {\bf a decay function}. This is an equation describing how the
intensity of the emotion decays over time, where time is given in
seconds. Feelings which do not decay (e.g. a permanent dislike of a
particular player) are not considered to be emotions for our
purposes, and belong among the static characteristics of the
character.

\end{itemize}

So, if a character is very sad about Team A having just scored, the
relevant emotional structure might be:

\begin{quote}

(type:sadness, intensity:10, target:nil, cause:(scored team:a time:125)
decay:1/t)

\end{quote}

The intensity of this emotion would go down each second for ten
seconds, then when it goes below one on the eleventh second, the
emotional structure would be deleted from the emotion pool.

An emotion structure generation rule consists of a set of
preconditions, which are to be filled by matching them on the
currently true facts about the world and about the character, and
currently active emotion structures, the emotional structures to be
added to the emotion pool, and the emotional structures to be
removed. For example:
\begin{quote}
\noindent{\bf Preconditions}:\\
\indent	(supports team: ?team)\\
\indent	(scores team: ?team)\\
\noindent{\bf Emotional structures to add}:\\
\indent	(type: happiness intensity: 8 target: nil 
\indent	cause: (scores team: ?team) decay: 1/t)\\
\noindent{\bf Emotional structures to delete}:\\
\indent	none
\end{quote}
This rule indicates that, if the team that the commentator supports
scores, a happiness structure should be added to the emotion pool.

There are only two ways for an emotion structure to be removed from
the emotion pool: it can be explicitly removed by an emotion structure
update rule, or its intensity can decay to below one, in which case it
is automatically removed. In future, it might be necessary to develop
a more sophisticated emotion maintenance system; however, since we
have no ambitions to cognitive plausibility, we will only add such
complications as necessary. We expect that this very simple emotional
maintenance method will suffice for most of the situations a soccer
commentator is likely to face.

Both emotion generation and behaviour generation are influenced by the
{\bf static characteristics} of the commentator character. This is a
set of static facts about the character, such as his nationality, the
team he supports, and so on. It is used to inform emotion and
behaviour generation, allowing a character to react in accordance with
his preferences and biases. For example, if a character supports the
team which is winning, his emotional state is likely to be quite
different that if he supports the losing team.

These emotion-generation rules can be arbitrarily complex, to take
into account the context of both the character's static
characteristics and the state of the world (in the case of soccer, the
game).

\subsection{Emotional behaviours}

Emotion structures and static characteristics are preconditions to the
activation of high-level emotion-expressing behaviours. These in turn
decompose into lower-level behaviours. The lowest level behaviours
specify how the text output by the text generation system is to be
marked up.

Emotionally-motivated behaviours are organized in a hierarchy of
mutually inconsistent groups. If two or more activated behaviours are
inconsistent, the one with the highest activation level is
performed. This will usually result in the strongest emotion being
expressed; however, a behaviour which is motivated by several
different emotions might win out over a behaviour motivated by one
strong emotion.

It is entirely possible for mixed emotional expressions to be
generated, as long as they are not inconsistent. For example, a happy
and excited character might express excitement by raising the pitch of
his voice and happiness by smiling. However, it is less likely that a
character will have a way to express, say, happiness and sadness in a
consistent manner.

\subsection{Text generation}

Character has a role to play in natural language generation. For
example, a character from a particular country or region might use the
dialect of that area, or a child character might use simpler
vocabulary and grammar than an adult. The current emotional state of
the character would also have an effect: an excited, angry character
might use stronger language and shorter sentences than a calm happy
one, for example. Loyall's work in the OZ project \cite{Loyall}
discusses some of these issues.

Despite these possibilities for affective generation, text generation
is an area in which Byrne does almost nothing of interest at
present. It is done very simply through a set of templates. Each
template has a set of preconditions which constrain the game
situations they can be used to describe. If more than one template
matches the chosen content, then the selection is based on how often
and how recently the templates have been used.

Byrne's text generation module does not generate plain text, but
rather text marked up with SEEML (see below). Although the speech
synthesis system we use can generate reasonable speech from plain
text, it is helpful to retain some phrase structure and part of speech
(POS) information from the natural language generation process to help
the speech sythesis system to generate appropriate prosody.

Moreover, linguistic information embedded in the text also helps
determine appropriate interruption points, should a more important
fact need to be expressed. Here we assume that Byrne should finish the
phrase it is currently uttering before it interrupts and starts a new
utterance. This is a very simplistic approach, and may not be
adequate.

Finally, linguistically-motivated facial gestures and speech
intonation are now hard-coded into the templates. If the natural
language generation system were more sophisticated, then a
post-generation gesture and intonation system might be necessary, but
with simple template generation this is the most effective
method.

\subsection{Expressive mark-up}
\label{markup}

SEEML\footnote{SEEML is pronounced ``seemly''.} (the speech, expression and
emotion mark-up language) is really just a slightly supplemented
superset of three different mark-up systems, namely FACSML, SABLE and
GDA. GDA is used to inform linguistically motivated expressive
behaviours, and also to aid the speech sythesizer in generating
appropriate prosody. FACSML is used to add facial behaviours, and
SABLE is used to control the speech synthesis.

The information encoded in SEEML is of five main types. 

\begin{itemize}

\item {\bf semantic/pragmatic structural information}: These tags
encode linguistic information in the text, so that prosody and
linguistically motivated facial gestures can be appropriately
generated. This information is mostly provided by the text generation
system, and is represented in GDA.

\item {\bf facial expressions}: These are combinations of FACS action
units into expressions of emotion. For example, {\bf smile} is a
facial expression. We use six facial expressions, based on Ekman's
basic emotions \cite{Ekman}. 

\item {\bf action units}: Low-level FACS action units can also be
marked into the text. This are mainly used for linguistically
motivated facial gestures, but are also used so that a particular
character's quirks of emotional expression can be integrated
smoothly. For example, a Mr Spock character might express doubt by
raising his eyebrow whenever Captain Kirk's name is mentioned. This
quirk could be expressed directly with AU4, the ``outer brow raiser''.

\item {\bf aural events}: These are short term aural events, and do
not have enclose text. For example, {\bf hiccup} is an aural event. 

\item {\bf {SABLE} tags}: Low-level SABLE tags can also be included in
the text. 

\end{itemize}

The expression mark-up module adds emotionally motivated mark-up to
the already marked-up text from the text generation system. Conflicts
are resolved in a simple (perhaps simplistic) manner. The combination
rules are:

\begin{itemize}

\item Unless identical, tags are assumed to be independent. Any
potential practical conflicts are left to be resolved by the speech
sythesizer and/or facial animation systems.

\item If two identical tags are assigned to a piece of text, the one
with the smaller scope is assumed to be redundant, and removed.

\item If two otherwise identical tags call for a change in some
parameter, it is assumed that that change is additive.

\end{itemize}

\subsection{{SEEML} verifier and style file}

The SEEML verifier interprets SEEML tags in the context of a style
file, adds time markers and lip synching information, and sends
appropriate FACS to the facial animation system and SABLE
(supplemented with phrase structure and POS tags) to the speech
sythesis system.

Although sophisticated lip-synchronization algorithms have been
developed (e.g. in \cite{Waters}), they are not necessary for our
purposes. Instead, we use a simple `cartoon style' lip animation,
which only shows the more obvious phoneme-viseme matches, as described
in \cite{Parke}.

The style file contains speech and animation system specific
interpretation rules. For example, it would determine the FACS which
are to be used to indicate a {\bf smile} for this particular face
model, and the sound file to be used for a {\bf hiccup}.

\section{Implementation}

In the first implementation, Byrne will use Franks and Takeuchi's
facial animation system \cite{Takeuchi} and the Festival speech system
\cite{Festival}. We hope that the standardized mark-up of the output
will allow the use of other systems as well.

The facial animation system is based on Waters' \cite{Waters} polygon
face model, and was implemented by Franks in C++ to run on a Silicon
Graphics machine. It has a wide range of expressions, although the lip
movement does not allow for sophisticated lip synching.

Festival is a concatenative speech sythesis system. It can generate
speech in a number of different languages and voices, and the
resulting speech is reasonably natural-sounding, although the user's
control of the speech synthesis is quite coarse. Its other important
feature for our purposes is its support of the Sable speech markup
system.

Byrne itself will be implemented in C++. We expect to have a
prototype, with the simplest emotional behaviours, ready for late June
1998.

\section{Future work}

Although the emotional behaviours outlined above are very simple,
this architecture allows for extremely complex behaviours to be
implemented. Maintaining consistency and believability in these
behaviours is expected to be a significant problem. We plan to develop
a set of tools for character designers to help them in in this task.

Also, the interruption mechanism described in section~\ref{choice} is
too simplistic, and is expected to result in very jumpy commentary. If we
find that this is the case, we will have to devise a more
sophisticated mechanism.

Although the character described throughout is a play-by-play
commentator character, we hope to develop other characters for the
soccer simulator domain, such as a coach or a soccer fan. We would
also like to develop a colour commentator~\footnote{A colour
commentator provides background details on teams and players,
such as statistics or amusing anecdotes.} to work with the
play-by-play, which would necessitate introducing a turn-taking
mechanism into the system.

Finally, the system described assumes pre-analysed facts about the
state of play in the soccer game. A high priority for us, therefore,
is to develop an analysis system which can output that kind of
information, so that Byrne can commentate a full simulated soccer game
in real time.

\section{Conclusion}

In this paper we have motivated and described an architecture for a
soccer commentator system, Byrne. Byrne will generate emotional,
expressive commentary of a RoboCup simulator league soccer game. It
makes heavy use of inter-system standards, so that the system can take
advantage of advances in speech synthesis, facial animation and
natural language generation.

We expect to have a working prototype completed by summer 1998.

\section{Acknowledgements}

We would like to thank Hiroaki Kitano for encouragement and support on
this project, and both Sean Luke and Ian Frank for their comments.

\nocite{Cahn}
\bibliographystyle{plain}
\bibliography{biblio}

\begin{thebibliography}{10}

\bibitem{Rocco}
Elisabeth Andre, Gerd Herzog, and Thomas Rist.
\newblock Generating multimedia presentations for robocup soccer games.
\newblock Technical report, {DFKI GmbH}, German Research Center for Artificial
  Intelligence, D-66123 Saarbrucken, Germany, 1998.

\bibitem{Festival}
Alan~W. Black, Paul Taylor, and Richard Caley.
\newblock {\em The Festival Speech Sythesis System}.
\newblock CSTR, University of Edinburgh, 1.2 edition, September 1997.

\bibitem{Blumberg}
Bruce Blumberg and Tinsley Galyean.
\newblock Multi-level control for animated autonomous agents: Do the right
  thing... oh, not that...
\newblock In Robert Trappl and Paolo Petta, editors, {\em Creating
  Personalities for Synthetic Actors}, pages 74--82. Springer-Verlag Lecture
  Notes in Artificial Intelligence, 1997.

\bibitem{Cahn}
Janet Cahn.
\newblock Generating expression in sythesized speech.
\newblock Master's thesis, Massachusetts Institute of Technology Media
  Laboratory, Boston, May 1989.

\bibitem{Ekman}
Paul Ekman and Erika~L. Rosenberg, editors.
\newblock {\em What the face reveals: Basic and applied studies of spontaneous
  expression using the facial action coding system}.
\newblock Oxford University Press, 1997.

\bibitem{Improv}
Athomas Goldberg.
\newblock {IMPROV}: A system for real-time animation of behavior-based
  interactive synthetic actors.
\newblock In Robert Trappl and Paolo Petta, editors, {\em Creating
  Personalities for Synthetic Actors}, pages 58--73. Springer-Verlag Lecture
  Notes in Artificial Intelligence, 1997.

\bibitem{SGML}
Charles Goldfarb.
\newblock {\em The SGML Handbook}.
\newblock Clarendon Press, 1991.

\bibitem{HayesRoth2}
Barbara Hayes-Roth, Robert van Gent, and Daniel Huber.
\newblock Acting in character.
\newblock In Robert Trappl and Paolo Petta, editors, {\em Creating
  Personalities for Synthetic Actors}, pages 92--112. Springer-Verlag Lecture
  Notes in Artificial Intelligence, 1997.

\bibitem{Johnstone}
Keith Johnstone.
\newblock {\em Impro}.
\newblock Routledge Theatre Arts Books, 1992.

\bibitem{JSML}
Java speech markup language specification [0.5 beta].
\newblock Technical report, Sun Microsystems, 1997.

\bibitem{RoboCup}
Hiroaki Kitano.
\newblock Robocup.
\newblock In Hiroaki Kitano, editor, {\em Proceedings of the IJCAI workshop on
  Entertainment and {AI/ALife}}, 1995.

\bibitem{Loyall}
A.~Bryan Loyall.
\newblock Some requirements and approaches for natural language in a believable
  agent.
\newblock In Robert Trappl and Paolo Petta, editors, {\em Creating
  Personalities for Synthetic Actors}, pages 113--119. Springer-Verlag Lecture
  Notes in Artificial Intelligence, 1997.

\bibitem{Nagao}
Katashi Nagao and Koiti Hasida.
\newblock Automatic text summarization based on the global document annotation.
\newblock Technical report, Sony Computer Science Laboratory, 1998.

\bibitem{SoccerServer}
Itsuki Noda.
\newblock {\em Soccer Server Manual Rev. 2.00}, May 1997.

\bibitem{Parke}
Frederic~I Parke and Keith Waters.
\newblock {\em Computer Facial Animation}.
\newblock A K Peters Ltd, Wellesley, MA, 1996.

\bibitem{Reilly}
W.~Scott~Neal Reilly.
\newblock {\em Believable social and emotional agents}.
\newblock PhD thesis, School of Computer Science, Carnegie Mellon University,
  May 1996.

\bibitem{Sable}
Draft specification for sable version 0.1.
\newblock Technical report, The Sable Consortium, 1998.

\bibitem{STML}
R.~Sproat, Paul Taylor, and Amy Isard.
\newblock A markup language for text-to-speech synthesis.
\newblock In {\em Proceedings of {EUROSPEECH}}, Rhodes, Greece, 1997.

\bibitem{Takeuchi}
Akikazu Takeuchi and Steven Franks.
\newblock A rapid face construction lab.
\newblock Technical Report SCSL-TR-92-010, Sony Computer Science Laboratory,
  Tokyo, Japan, May 1992.

\bibitem{MIKE}
Kumiko Tanaka-Ishii, Itsuki Noda, Ian Frank, Hideyuki Nakashima, Koiti Hasida,
  and Hitoshi Matsubara.
\newblock {MIKE}: An automatic commentary system for soccer.
\newblock Technical Report TR-97-29, Electrotechnical Laboratory, Machine
  Inference Group, Tsukuba, Japan, 1997.

\bibitem{SSML}
Paul Taylor and Amy Isard.
\newblock {SSML}: A speech synthesis markup language.
\newblock {\em Speech communication}, 1997.

\bibitem{Waters}
Keith Waters and Thomas~M. Levergood.
\newblock {DECFace}: An automatic lip-synchronization algorithm for sythetic
  faces.
\newblock Technical Report {CRL 93/4}, Digital Cambridge Research Laboratory,
  September 1993.

\end{thebibliography}

\end{document}